\documentclass[prodmode,acmtap]{ci2017}
\usepackage{url}
\usepackage{soul}

\acmVolume{2}
\acmNumber{3}
\acmArticle{1}
\articleSeq{1}
\acmYear{2010}
\acmMonth{5}

\usepackage[ruled]{algorithm2e}
\SetAlFnt{\algofont}
\SetAlCapFnt{\algofont}
\SetAlCapNameFnt{\algofont}
\SetAlCapHSkip{0pt}
\IncMargin{-\parindent}
\renewcommand{\algorithmcfname}{ALGORITHM}

\graphicspath{{figs/}{figsOld/}}  

\usepackage{tikz}
\makeatletter
\newcommand*{\inputGraphics}[1]{%
  \ifx\input@path\@undefined
    \def\reserved@a{%
      \let\input@path\Ginput@path
      \InputIfFileExists{#1}{}{file not found}%
      \let\input@path\@undefined
    }%
  \else
    \edef\reserved@a{%
      \noexpand\let\noexpand\input@path\noexpand\Ginput@path
      \noexpand\InputIfFileExists{#1}{}{file not found}%
      \noexpand\def\noexpand\input@path\expandafter{\input@path}%
    }%
  \fi
  \reserved@a
}
\makeatother

\usepackage{floatrow}
\usepackage{subfig}
\usepackage{caption}
\usepackage{color}
\definecolor{Green}{RGB}{10,200,100}
\definecolor{Blue}{RGB}{0,0,255}

\title{Diversity of preferences can increase collective welfare in sequential exploration problems}
\author{PANTELIS P. ANALYTIS \affil{Cornell University}
HRVOJE STOJIC \affil{University College London}
ALEXANDROS GELASTOPOULOS \affil{Boston University}
MEHDI MOUSSA\"{I}D \affil{Max Planck Institute for Human Development}
}

\begin{abstract}
Diversity has been found beneficial for groups, as complementarities help their members solve complex problems and the group members can leverage the wealth of information possessed by different individuals. Could diversity be beneficial in problems where groups of people sequentially explore large numbers of alternatives, such as in search engines, marketplaces, and other online interfaces?  We use a search model to study the problem, which exemplifies the trade-off between exploiting the best alternative found thus far and exploring further with the hope of identifying better alternatives. In our model, agents with diverse yet correlated preferences search the alternatives in order of popularity and choose the first alternative with utility higher than a certain satisficing threshold. We find that some diversity of preferences has a beneficial effect for collective welfare, especially in environments with high search costs. Although diverse collectives pay a higher search cost, this cost can be outweighed by finding alternatives with higher utility. 

\end{abstract}

\begin{document}

\maketitle


\section{Introduction}



In search engines, online marketplaces and other human--computer interfaces large collectives of individuals sequentially interact with numerous alternatives of varying quality. In these contexts, individual trial and error (exploration) is crucial for uncovering novel high-quality items or solutions, but entails a high cost for individual agents \cite{frazier2014incentivizing}. Self-interested decision makers, we will show, are often better off imitating the choices of individuals who have already incurred the costs of exploration. Although imitation makes sense at the individual level, it deprives the group of additional information that could have been gleaned by individual explorers \cite{Rogers1988does}. Under these grim circumstances, certain non-monetary mechanisms can keep imitation forces in check and allow the collective to reap some of the benefits of the independent collection of information. For example, in simultaneous exploration problems, a natural equilibrium evolves between explorers and imitators \cite{Conlisk1980costly,Kameda2002cost}. Further, in some collective exploration settings, barriers to communication such as a sparser communication network among individuals can prove beneficial at the collective level. They encourage people to explore more, thus supplying useful information to the group \cite{Fang2010balancing,Lazer2007network,Mason2008propagation,toyokawa2014human}. 

Diversity is known to be a blessing for groups and collectives, as they can leverage the wealth of information possessed by different individuals \cite{Conradt2013swarm,Davis2014crowd,Muller2017wisdom} or take advantage of the complementarities between group members to solve complex problems \cite{Clearwater1991cooperative,Hong2004groups}. Could some preference diversity be beneficial in problems where collectives sequentially explore numerous alternatives, and thus despite reducing the immediate value of social learning lead to an increase in collective welfare?  


%
%




\section{The model}

We use a search model to study the problem, as it exemplifies the trade-off between exploration and exploitation \cite{march1991exploration}. We assume that there are $N$ alternatives (or products) $X_1,...,X_N$ in a market populated by $M$ agents $A_1,...,A_M$. The alternatives have an objective utility component $u_{o}$, which is identical for all agents, and a subjective component $u_{s}$,  which is agent specific. The objective component $u_{o}$ of each alternative is a draw from an independent and identically distributed (iid) random variable, normally distributed with mean $\mu_{o}$ and variance $\sigma^2_{o}$. The subjective component $u_{s}$ of each alternative is iid normal with mean $\mu_{s}$ and variance $\sigma^2_{s}$. The overall utility of an alternative $n$ is then a sum of two draws from iid normal variables, $u_{n} =  u_{no} + u_{ns}$, which itself is an iid normal variable with variance $\sigma^2 = \sigma^2_{o} + \sigma^2_{s}$. 

    
The agents encounter alternatives sequentially and can learn the utility of an alternative $u_n$ only by sampling it and paying a search cost $c$. The agents can sample as many alternatives as desired, but they can choose only one of the alternatives they have sampled. An example of sampling behavior would be skimming through book recommendations at Amazon or examining the results provided by a search engine. We assume that the search cost is constant and that there is no post-sampling uncertainty---by sampling the alternative, the agents learn its true utility. An agent's return depends on the utility of the best alternative discovered so far and the search cost, $ max(u_1,u_2,...u_n) - n \times c $. The returns from sampling one more alternative can be formulated as:
\begin{eqnarray}
         &   & E(u_n - max(u_1,u_2,...u_{n-1}) | u_n > max(u_1,u_2,...u_{n-1})) - c \nonumber
\end{eqnarray}
When viewed on the level of a single agent, this is a classical optimal stopping problem---as studied extensively in statistics and economics \cite{Degroot1970optimal,analytis2014collective}---in which after examining each new alternative the agent decides whether to sample further or to stop search. For random search, the problem has an optimal solution, which can be expressed in the form of a stopping threshold. 


In our model, agents do not search randomly. Instead, they follow public popularity information generated by choices made by previous searchers in the market. After the first agent, a second or $m^{th}$ agent is asked to make its choice. Now the agent can observe the sum of choices made by previous searchers in the form of a popularity vector $P= \{P_1,...,P_N\}$ that records the choices of alternatives $X_1,...,X_N$ and is updated whenever an individual makes a choice. Examples of such popularity information would be product sales, number of song downloads, or number of article citations. The agents sample the alternatives $ X_1,X_2...X_n $ in decreasing order of popularity $P_1 \geq P_2 \geq... \geq P_n$, where popularity is defined as the number of times that an alternative has been selected in the past. When there is a tie, the agents choose which one to sample next at random.  We call the order in which the alternatives are searched the \emph{search path}. The agents stop search when they encounter an alternative $u_n$ with utility higher than a threshold $T$; otherwise, they continue sampling until the alternatives are exhausted. The proposed model captures the behavioral regularities of search in online interfaces \cite{joachims2005accurately,craswell2008experimental} and bolsters them with a fully fleshed-out utility framework.



\section{Simulations and results}

  


We simulated markets consisting of 1000 agents that decided sequentially which of 100 alternatives to choose, after searching the market according to public popularity information. We systematically varied the diversity of preferences in the market, $d$, by setting the subjective variance of each individual agent $\sigma^2_{s}$ equal to $d \in \{0, 0.1,...,0.9,1 \} $ and the variance of the objective utility $\sigma^2_{o} $ equal to $ 1 - d $. We drew subjective utilities for each agent independently, whereas the draw of the objective utilities was the same for all agents. 
Next, we varied the cost of search, $ c $, that agents have to pay when sampling an alternative, $c  = \{ 1/2^2, 1/2^3,1/2^4, 1/2^5, 1/2^6 , 1/2^7, 1/2^8 \}$. These costs correspond to the threshold parameters $ T = \{ 0.34, 0.78 , 1.15 , 1.47, 1.76 , 2.03, 2.28 \} $; these parameters are optimal for a random search model \cite{chow1971great}, which our agents used despite searching according to the popularity ranking. This yields $11 \times 7 = 77$ markets, each of which was simulated 1000 times to obtain stable averages.%
\footnote{The code for the simulations can be openly accessed at https://osf.io/up5pe/}

\begin{figure*}
\centering
\includegraphics[width=1\textwidth]{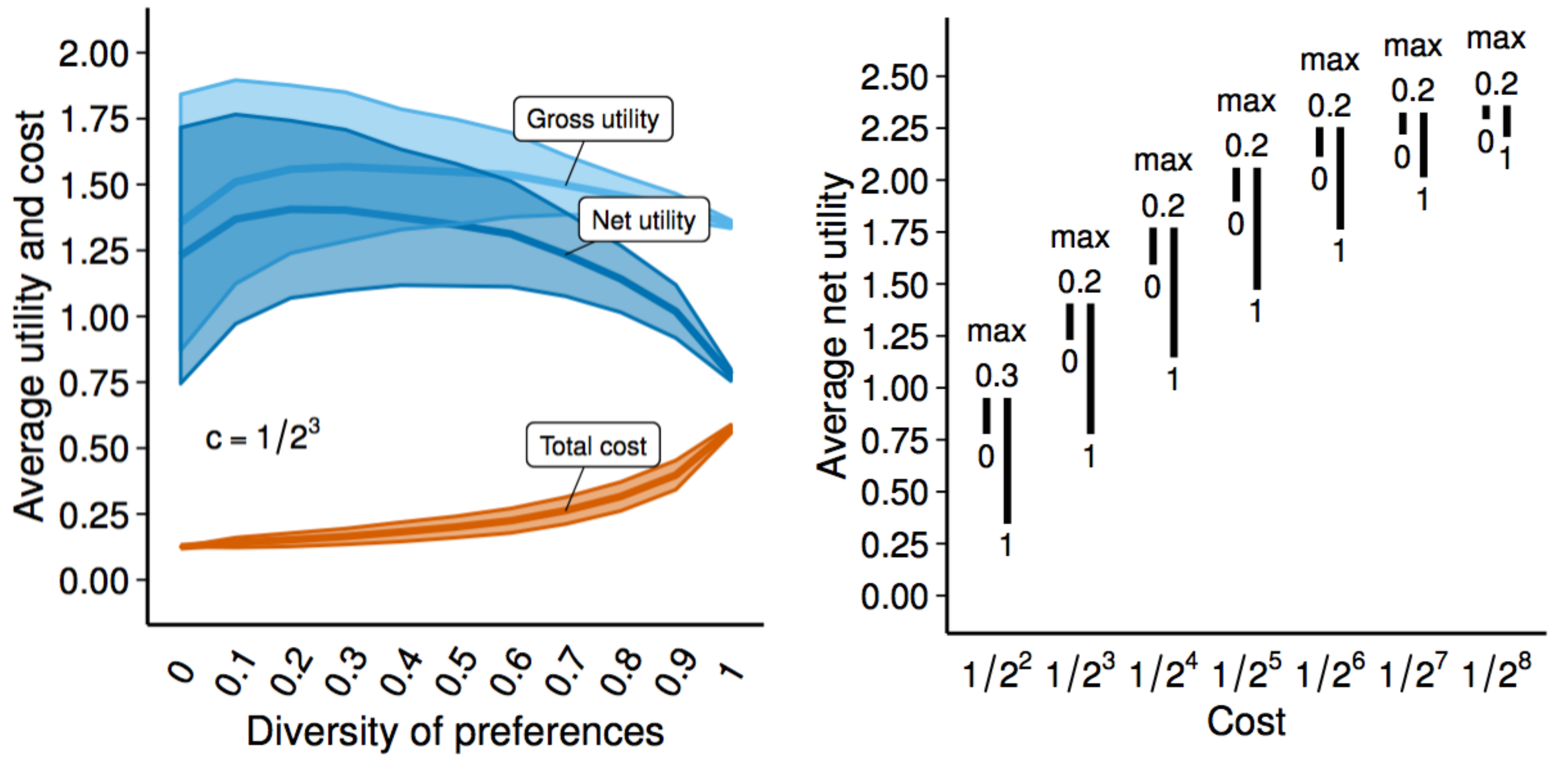}
\vspace{-6mm}
\caption{\small
 The average net utility in the market is highest at intermediate levels of preference diversity. The lines represent the average gross and net utility of the selected alternative and the average total cost of search at a search cost of $1/2^3$. The ribbons represent the variability in 1000 repetitions of the simulation. The beneficial effect of diversity increases with the search cost. For each cost level, we illustrate the difference between the level of diversity with the maximum net utility and conditions with no diversity, $d=0$, and maximum diversity, $d=1$.%
} 
\label{fig_welfare_all}
\end{figure*}



In Figure 1 (left), we illustrate the average utility and total cost for the agents in the market, focusing on the second higher search cost case $c = 1/2^3$. In the maximum diversity case, $d = 1$, the environment is recreated randomly for each agent. In this case, the popularity information is completely uninformative and welfare outcomes are equal to those in the market where agents search randomly. At $d = 1$, agents incur the highest average total costs and earn the lowest average net utility from the market, even though the thresholds they employ are optimally set. At $d = 0$, the alternative chosen by the first agent satisfies all the subsequent agents in the population, leading to herding. In these markets, the cost of search is minimal, and the obtained net utility is much higher than in the random search market. Because the first agent may settle on a very good or a simply passable alternative, $d=0$ implies more variability in the average utility in the market. This is illustrated by the ribbons in Figure 1 (left). Although the search costs are minimized, the economy does not maximize the collective welfare of the agents involved. There is a robust non-monotonic relationship between $d$ and the average net utility---the highest level of utility is consistently achieved for $d \approx 0.2$, whereas at high costs even markets with $d = 0.5$ on average do better than markets with no diversity whatsoever. Although the average agent incurs a higher average cost of search in markets with $d > 0$ than in a market with $d = 0$, this loss is outweighed by the benefits of choosing alternatives with higher gross utility. Figure 1 (right) illustrates that the non-monotonic relation between diversity and collective utility persists even at much lower costs of search, although the average differences between conditions decrease.



What is the mechanism that leads to an increase in average net utility in markets with some diversity of preferences? In a market with no diversity whatsoever, everybody herds on the option selected by the first individual. Occasionally, this turns out to be an excellent alternative. More often than not, however, it is located just off the satisficing threshold. The collective welfare is initial outcome-dependent \cite{page2006path}. In contrast, in markets with some diversity of preferences, agents do not always appreciate the choices of previous individuals and they venture further until they encounter a satisficing option. With each choice, the \emph{search path} followed by the agents improves. Eventually, alternatives with higher objective utility components are more likely to occupy the first positions in the \emph{search path}. Thus, agents deciding later on in time possess a useful popularity signal, require relatively few steps to find good alternatives, and tend to settle on better outcomes (Figure 1, gross utility). In essence, some diversity of preferences allows the agents to benefit from the thick informational externalities that are inherent in sequential social interactions. By searching and deciding, the agents unintentionally provide useful information to the individuals deciding after them.

\newpage
\bibliographystyle{ci-format}
\bibliography{library,socialLearning}

\end{document}